\documentclass{article} % For LaTeX2e
\usepackage[pagebackref=true,breaklinks=true,letterpaper=true,colorlinks,bookmarks=false]{hyperref}
\usepackage{iclr2015,times}
\usepackage{hyperref}
\usepackage{url}

\usepackage{amsmath,amssymb}
\usepackage{amsthm}
\usepackage{graphicx}
\usepackage[tight,scriptsize]{subfigure}
\usepackage{enumitem}

\def\real{\mathbb{R}}
\def\eg{{\em e.g., }}

\def\cut#1{{\color{red}}}
\def\cutForReview#1{{\color{blue}}}

\def\cutCNN#1{{\color{green}}}

\newcommand{\be}{\begin{equation}}
\newcommand{\ee}{\end{equation}}
\newcommand{\bea}{\begin{eqnarray}}
\newcommand{\eea}{\end{eqnarray}}

\newcommand{\vspacetitle}{\vspace{-0cm}}

\newtheorem{rem}{Remark} 
\newtheorem{thm}{Theorem}

\newtheorem{cor}{Corollary} 

\def\xii{{I(x_i)}} 
\def\xx{{I}}
\def\M{{\xi}}

\def\xii{{y_i}} 
\def\xx{{\bf y}}
\def\M{{\bf z}}

\def\ttheta{{\alpha}}
\def\rrho{{\rho}}

\def\xii{{x_i}} 
\def\xx{{x}}
\def\M{{\theta}} % CAREFUL IT INTERFERES WITH \ttheta, gradient orientation

\def\f{{y}}

\title{Visual Scene Representations: \\ {\large Scaling and Occlusion in Convolutional Architectures\thanks{Also UCLA Technical Report CSD140024, November 12, 2014}}} 

\author{
Stefano Soatto, Jingming Dong \& Nikolaos Karianakis\\
UCLA Vision Lab\\
University of California, Los Angeles\\
Los Angeles, CA 90095, USA\\
\texttt{\{soatto,dong\}@cs.ucla.edu, nikarianakis@ucla.edu}
}

\iclrfinalcopy % Uncomment for camera-ready version

\begin{document}

\maketitle

\begin{abstract} 
We study the structure of representations, defined as approximations of minimal sufficient statistics that are maximal invariants to nuisance factors, for visual data subject to scaling and occlusion of line-of-sight. We derive analytical expressions for such representations and show that, under certain restrictive assumptions, they are related to features commonly in use in the computer vision community. This link highlights the conditions tacitly assumed by these descriptors, and also suggests ways to improve and generalize them. 
\end{abstract}

\section{Introduction}
\vspacetitle

\cut{A visual representation is a function of past images that is useful to answer questions about the scene given future images from it, regardless of nuisance variability that will affect them.} \cite{soattoC14ICLR} define an optimal representation  as a minimal sufficient statistic (of past data for the scene) and a maximal invariant (of future data to nuisance factors), and propose a measure of how ``useful'' (informative) a representation is, via the uncertainty of the prediction density. What is a nuisance depends on the task, that includes decision and  control actions about the surrounding environment, or {\em scene}, and its geometry (shape, pose), photometry (reflectance), dynamics (motion) and semantics (identities, relations of ``objects'' within).\cut{ Depending on the task, nuisance variables may include viewpoint, illumination, sensor calibration, and {\em occlusion} of line of sight. In this paper we focus on the latter and its impact in the design and learning of representations. }

%%%%%%%%%%%%%%%%%%%%%%%%%%%%%%%%%%%%%%%%%%%%%%%%%%
%									   Related Work
%%%%%%%%%%%%%%%%%%%%%%%%%%%%%%%%%%%%%%%%%%%%%%%%%%
\cut{\vspacetitle
\subsection{Related Work and Contributions}
\vspacetitle
}
\cut{This paper builds on \cite{soattoC14ICLR} by focusing on occlusion and scaling phenomena. There, a representation is characterized as an approximation of the likelihood function, with nuisance factors either marginalized and profiled ({\em max-out}). Most work in {\em low-level vision} handles occlusions by restricting the attention to local regions of the {\em image}, resulting in representations known as {\em local descriptors} -- too many to review here, with SIFT a prototypical representative \cite{lowe04distinctive}. Scale changes are handled by performing computation in {\em scale-space} \cite{lindeberg98}. Empirical comparisons abound (\eg \cite{mikolajczyk04comparison}) and recently expanded to include convolutional networks \cite{fischer2014descriptor}. Our work is aimed at understanding how to relate various descriptors to each other, so the assumptions on which they rely become patent, and to an ``ideal'' representation, so one can see how to improve them, not just compare them on any given dataset.
}
We show that optimal management of nuisance variability due to occlusion is generally intractable, but can be approximated leading to a composite (correspondence) hypothesis test, which provides grounding for the use of ``patches'' or ``receptive fields,'' ubiquitous in practice\cut{ (Sect. \ref{sect-receptive})}. The analysis reveals that the size of the domain of the filters should be {\em decoupled} from spectral characteristics of the image, unlike traditionally taught in scale-space theory, an unintuitive consequence of the analysis. This idea has been exploited by \cite{dongS15} to approximate the optimal descriptor {\em of a single image}, under an explicit model of image formation (the Lambert-Ambient, or LA, model) and nuisance variability, leading to DSP-SIFT. Extensions to multiple training images, leading to MV-HoG and R-HoG, have been championed by \cite{dongHBSB13}. Here, we apply domain-size pooling to the scattering transform \cite{mallatB11} leading to DSP-SC, to a convolutional neural network, leading to DSP-CNN, and to deformable part models \cite{felzenswalb}, leading to DSP-DPM, in Sect. \ref{sect-dsp-sca}, \ref{sect-DSP-CNN} and \ref{sect-DSP-DPM} respectively.

\cut{\vspacetitle
\section{Background} 
\vspacetitle
}
We treat images as random vectors $\xx, \f$ and the scene $\M$ as an (infinite-dimensional) parameter. An optimal representation is a function $\phi$ of past images $\xx^t \doteq \{\xx_1, \dots, \xx_t\}$ that maximally reduces uncertainty on questions about the scene \cite{geman2015visual} given images from it and regardless of nuisance variables $g\in G$. In \cite{soattoC14ICLR} the sampled orbit anti-aliased (SOA) likelihood is introduced as: 
\be
\hat L_{G,\epsilon} (\M; \xx) = \max_{i} \hat L(\M, g_i; \xx), \ i = 1, \dots, N(\epsilon)
\ee
where 
\be
\hat L(\M, g_i; \xx) \doteq \int_G L(\M, g_i g; \xx) dP(g)
\ee
and $L(\M, g; \xx) \doteq p_{\M,g}(\xx)$ is the joint likelihood, understood as a function of the parameter $\M$ and nuisance $g$ for fixed data $\xx$, with $dP(g) = w(g^{-1})d\mu(g)$ an {\em anti-aliasing} measure with positive weights $w$. The SOA likelihood is an optimal representation in the sense that, for any $\epsilon$, it is possible to choose $N$ and a finite number of samples $\{g_i\}_{i=1}^N$ so that $\phi_\M(\xx^t) \doteq \hat L_{G, \epsilon}(\M; \xx^t)$ approximates to within $\epsilon$ a minimal sufficient statistic (of $\xx^t$ for $\M$) that is maximally invariant to group transformations in $G$. This result is valid under the assumptions of the Lambert-Ambient (LA) model \cite{soattoD12ICVSS}, which is the simplest known to capture the phenomenology of image formation including scaling, occlusion, and rudimentary illumination.\cut{ For us, what matters of the LA model are three facts: First, the scene {\em separates}  the past from the future: $\xx^t \perp \f \ | \ \M$, meaning that $p_\M(\xx^t, \f) = p_\M(\xx^t)p_\M(\f)$.  Second, conditioning on viewpoint factorizes the likelihood: If $g\in G = SE(3)$ is the position and orientation of the camera in the reference frame of the scene $\M$ and the image $\f$ is made of pixels $\f_i$, then
\be
\vspace{-.3cm}
p_\M(\f | g) = \prod_{i} p_\M(\f_i | g)
\label{eq-prod}
\ee
Third, the action of restricted groups $G\subset SE(3)$, for instance planar translations, rotations, scalings, affine and projective transformations, contrast transformations, etc. is {\em approximately} equivariant, in the sense that for a sufficiently small domain, 
\be
p_\M(g_1\f | \bar g_2) = p_\M(\f | \bar g_1 \bar g_2)
\label{equi}
\ee
where the product $g_1 g_2$ denotes group composition and the bar (omitted henceforth) denotes the embedding of the group action on the (2-D) plane into (3-D) Euclidean space.  In Sect. \ref{sect-occlusion}  we will motivate these assumptions by restricting the representation to local spatial domains, and use it in Sect. \ref{sect-viewpoint} to achieve invariance to arbitrary vantage points. When the task corresponds to a partition of the space of scenes $\M$, for instance those providing the same answer to a finite collection of questions based on (future) data $\f$ and represented by a (supervised, past) training set $\xx^t$, then $\phi_{\M, G}(\f) \doteq \phi_{\M, G}(y) \phi_\M(\xx^t) %\simeq \hat p_{{}_{X^t, G}}(\f)\hat p_{{}_{X^t}}(\xx^t) 
\propto \hat p_{{}_{x^t, G}}(\f)$ can be considered a ``learned approximation'' of an optimal representation. We now illustrate how to compute such an approximation explicitly under the assumptions of the LA model.
}
\cut{\begin{rem}[What you lose if you use lousy view(s) -- Active Sensing]
A representation, informative as it may be, can be no more informative than the data itself, uninformative as it may be. This is irrelevant in our context, for we are seeking statistics that are  {\em as informative as the (training) data} (sufficient), however good or bad that is. For the representation to (asymptotically) approach the informative content of the {\em scene,} it is necessary to {\em design the experiment} so that the data collected $\xx^t$, with $t \rightarrow \infty$, yields statistics that are asymptotically {\em complete} \cite{fedorov1972theory}. Such active learning or active sensing is beyond the scope of this paper.
\end{rem}
When a single training datum is given, $\xx^t = \xx$, no intrinsic (intra-class) variability can be learned, and the variability in the data is ascribed to the nuisances. The representation for $t = 1$ thus reduces to 
\be
\phi_{\xx,G}(\f) = p_{G}(\f | \xx)
\label{ideal-one}
\ee
which is approximated locally by DSP-SIFT \cite{dongS15}.
}

\vspacetitle
\section{Constructing Visual Representations} 
\label{sect-nuisances} 
\vspacetitle

\cut{In \cite{soattoC14ICLR}, it is shown that the orbit likelihood of the LA model is maximally invariant and minimally sufficient. Thus, visual representations can be trained or designed to compute the SOA likelihood with respect to nuisances that include illumination, viewpoint (with the associated scale changes), and partial occlusions.}

\cut{\vspacetitle
\subsection{Contrast invariance} 
}\label{sect-contrast}
\vspacetitle

\cut{Contrast transformations are monotonic continuous transformation of the (range space of the) data. If applied globally to an image, they are a crude approximation of changes in the image due to illumination. However, applied locally and independently in each receptive field, they can capture complex illumination effects. As we will see, occlusion will force our representation to be restricted to local statistics, so we adopt local contrast transformation as a model of illumination changes. It is well-known that the curvature of the level sets \cutForReview{at each point} is a maximal invariant \cite{alvarezGLM93}. Since the gradient orientation is everywhere orthogonal to the level sets, it is also a maximal contrast invariant.  The following expression for the invariant is obtained via marginalization of the norm of the gradient for a single training image, since the action of contrast is independent at each pixel.} 
\begin{thm}[Contrast invariant]
\label{claim-contrast}
Given a training image $\xx$ and a test image $\f$, assuming that the latter is affected by noise that is independent in the gradient direction and magnitude, then the maximal invariant of $\f$ to the group $G$ of contrast transformations  is given by 
\be
p_{{}_{\xx,G}}(\f)  = p(\angle \nabla \f | \xx) ~ \| \nabla \xx \|.
\label{eq-contr-inv}
\ee
\end{thm}
\cut{The independence assumption above is equivalent to assuming that the gradient magnitude and orientation of $\f$ are related to the gradient magnitude and orientation of $\xx$ by a simple additive model: $\| \nabla \f \| = \| \nabla \xx \| + n_\rrho$ and $\angle \nabla \f = \angle \nabla \xx \oplus  n_\ttheta$, where $\oplus$ denotes addition modulo $2\pi$, and $n_\rrho$ and $n_\ttheta$ are independent.
These are all modeling assumptions, clearly not strictly satisfied in practice, but reasonable first-order approximations.}
Note that, other than for the gradient, the computations above can be performed point-wise under the assumption of LA model, so we could write \eqref{eq-contr-inv} at each pixel $\f_i$: if  $\ttheta \doteq \angle \nabla \f_i$, 
\be
{\phi_\xx(\ttheta) 
= \prod_i {\cal N}_{{\mathbb S}^1}(\ttheta_i - \angle \nabla \xx_i; \epsilon_\ttheta)  \| \nabla \xx_i \|}
\label{eq-contr-inv2}
\ee
%\cut{In the rest of the paper,  we use the symbol $\ttheta$ to denote the orientation of the image gradient relative to one of the coordinate axes, and omit the subscript $G$ when referring to contrast (since the use of the argument $\alpha$ makes it unambiguous). The width of the kernel $\epsilon_\ttheta$ is a design (regularization) parameter.}
%
%\cut{\begin{rem}[No invariance for $\xx$]}
Note that \eqref{eq-contr-inv2} is invariant to contrast transformations of $\f$, but {\em not} of $\xx$.\cut{ For a single training image, the latter can be handled by {\em normalization} as we will see next. For multiple images, the factor can in principle be different for each training image. 
\end{rem}}
\cut{\begin{rem}[Bayesian invariant]
\label{rem-clamp}
In the proof of Theorem \ref{claim-contrast}, the gradient magnitude is marginalized with respect to the base measure. With a different prior, for instance arising from bounds on the gradient or from statistics of natural images, marginalization yields a factor other than $\| \nabla \xx \|$. Clamping, described next, can be understood as a particular choice of prior for marginalization of the gradient magnitude.
\end{rem}}
Invariance to contrast transformations in the (single) {\em training} image can be performed by normalizing the likelihood, which in turn can be done\cut{ in a number of ways. If contrast transformations are globally affine\cutForReview{ (they transform the intensity of each pixel by adding an offset and multiplying by the same scalar)}, then the joint likelihood can be normalized} by simply dividing by the integral over $\ttheta$, which is the $\ell^1$ norm of the histogram across the entire image/patch
\be
 \frac{\phi_\xx(\ttheta)}{ \| \phi_\xx(\ttheta) \|_{\ell^1}}  = \frac{p(\ttheta | \xx)  \| \nabla \xx\|}{\int p(\ttheta | \xx)d\ttheta   \| \nabla \xx\| } %=  \frac{p_{G}(\ttheta | \xx)  \| \nabla \xx\|}{  \| \nabla \xx\| } = p(\ttheta | \xx) 
= p(\ttheta | \xx)
\label{eq-normalize}
\ee
that should be used instead of the customary $\ell^2$ \cite{lowe04distinctive}. \cut{If the contrast transformation is non-linear, it cannot be eliminated by global normalization\cutForReview{, as the factor $\| \nabla \xx_{(u,v)}\|$ is potentially different at each location $(u,v)$}. }
\cut{\begin{rem}[Clamping]
\label{rem-clamping}
When the joint distribution is approximated by the product of marginals, as in \cite{lowe04distinctive}, joint normalization is still favored in practice as it introduces some correlations among marginal histograms \cite{dalalT05}. However, cells with large gradients tend to dominate the histogram, pushing all other peaks lower. Alternatively, one could independently normalize each of the histograms, $\phi_{\xii}(\ttheta)$  and then concatenate them. But this has the opposite effect: Cells with faint peaks, once re-normalized, are  given undue importance and relative intensity difference between different cells are discarded. A common trick consisting of joint normalization (so faint cells do not prevail) followed by ``clamping'' (saturation of the maximum to a fraction of the value of the highest peak, so large gradients do not dominate), and then re-normalization, seems to achieve a tradeoff between the two \cite{lowe04distinctive}. %\footnote{This process is justified there as follows: {\em ``non-linear illumination changes can also occur due to camera saturation or due to illumination changes that affect 3D surfaces with differing orientations by different amounts. These effects can cause a large change in relative magnitudes for some gradients, but are less likely to affect the gradient orientations. Therefore, we reduce the influence of large gradient magnitudes by thresholding the values in the unit feature vector to each be no larger than 0.2, and then renormalizing to unit length. This means that matching the magnitudes for large gradients is no longer as important, and that the distribution of orientations has greater emphasis.''} It is important to note, however, that this is only useful because we have replaced the joint likelihood with the product of marginals, and would not be needed otherwise.} 
This process can also be understood as a way of marginalizing contrast with respect to a different prior. 
\end{rem}
}
Once invariance to contrast transformations is achieved, which can be done on a single image $\xx$, we are left with nuisances $G$ that include general viewpoint changes, including the occlusions they induce. This can be handled by computing the SOA likelihood with respect to $G$ of $SE(3)$ (Sect. \ref{sect-viewpoint}) from a training sample $\xx^t$, leading to
\be
\hat L(\M, g_i; \xx^t) = \Big\{\int_G \phi_{\xx^t}(\alpha | g_i \circ g)dP(g) \Big\}_{i=1}^N
\label{soa-alpha}
\ee
%\cut{In the next section we show how to handle occlusions, and in the following one general viewpoint changes.
%\vspacetitle
%\subsection{Occlusions}
%\label{sect-occlusion} 
%\vspacetitle
%}
%\cut{We do not know ahead of time what portion of an object or scene, seen in training images, will be visible in a test image.}
Occlusion, or visibility, is arguably the single most critical aspect of visual representations. It enforces {\em locality}, as dealing with occlusion nuisances entails searching through, or marginalizing, all possible (multiply-connected) subsets of the test image. This  power set is clearly intractable even for very small images.
{\em Missed detections} (treating a co-visible pixel as occluded) and {\em false alarms} (treating an occluded pixel as visible) have different costs: Omitting a co-visible pixel from $\Omega$ decreases the likelihood by a factor corresponding to multiplication by a Gaussian for samples drawn from the same distribution; vice-versa,  including a pixel from $\Omega^c$ (false alarm) decreases the log-likelihood by a factor equal to multiplying by a Gaussian evaluated at points drawn from another distribution, such as uniform. 
So, testing for correspondence on {\em subsets of the co-visible regions}, assuming the region is sufficiently large, reduces the power, but not the validity, of the test. This observation can be used to {\em fix the shape} of the regions, {\em leaving only their size to be marginalized, or searched over.}\cut{\footnote{Alternatively, the sampling can be framed as a sequential hypothesis test for joint matching and domain size estimation, as in region-growing or quickest setpoint change detection\cutForReview{ \cite{vedaldiS06CVPR,sundaramoorthiSY10}}.}
}\cutForReview{ A region that is included in the ``true'' region will be accepted even if its likelihood is slightly lower than the full region. A region that straddles the occluding boundary, and therefore includes occluded regions, will be rejected as a whole. Note that this test must be performed for many regions, including different locations {\em and sizes}.} This reasoning justifies the use of ``patches'' or ``receptive fields'' to seed image matching, but emphasizes that a search over different {\em sizes} \cite{dongS15} is needed.
\cutForReview{
\begin{rem}[Visibility and the locality of representation]
Occlusion is what forces the representation to be {\em local}: We do not know what portion of a scene will be visible in a test image, and therefore we have to {\em search} across all subsets of the image. Once this is established, it facilitates consideration of viewpoint as a nuisance, since the restriction of viewpoint-induced deformation to a small (co-visible) subset of the image can be well approximated by a group transformation (projective, affine, or even similarity). Of course, the transformations have to be compatible with an overall rigid scene, is easily tested with the tools of epipolar geometry \cite{maSKS}. 
\end{rem}}

\cut{
\begin{figure}[tb]
\begin{center}
\includegraphics[width=.25\textwidth]{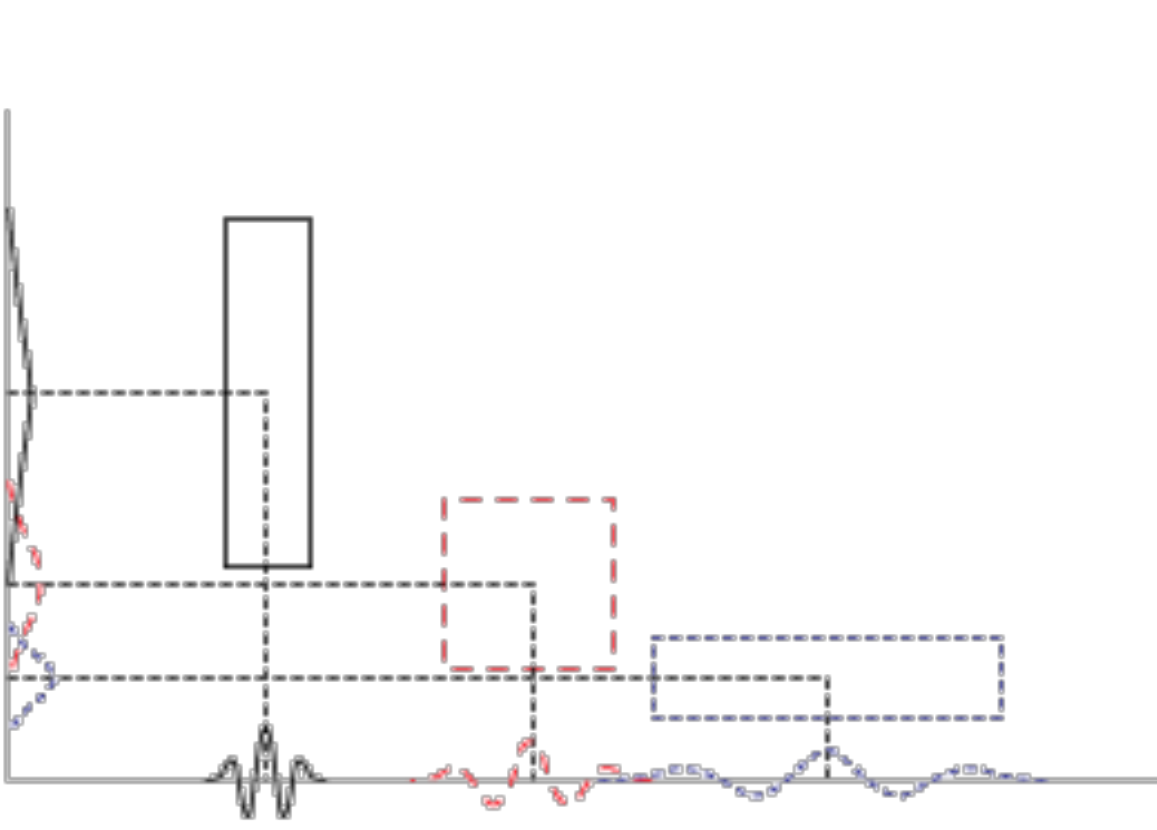}
\includegraphics[width=.25\textwidth]{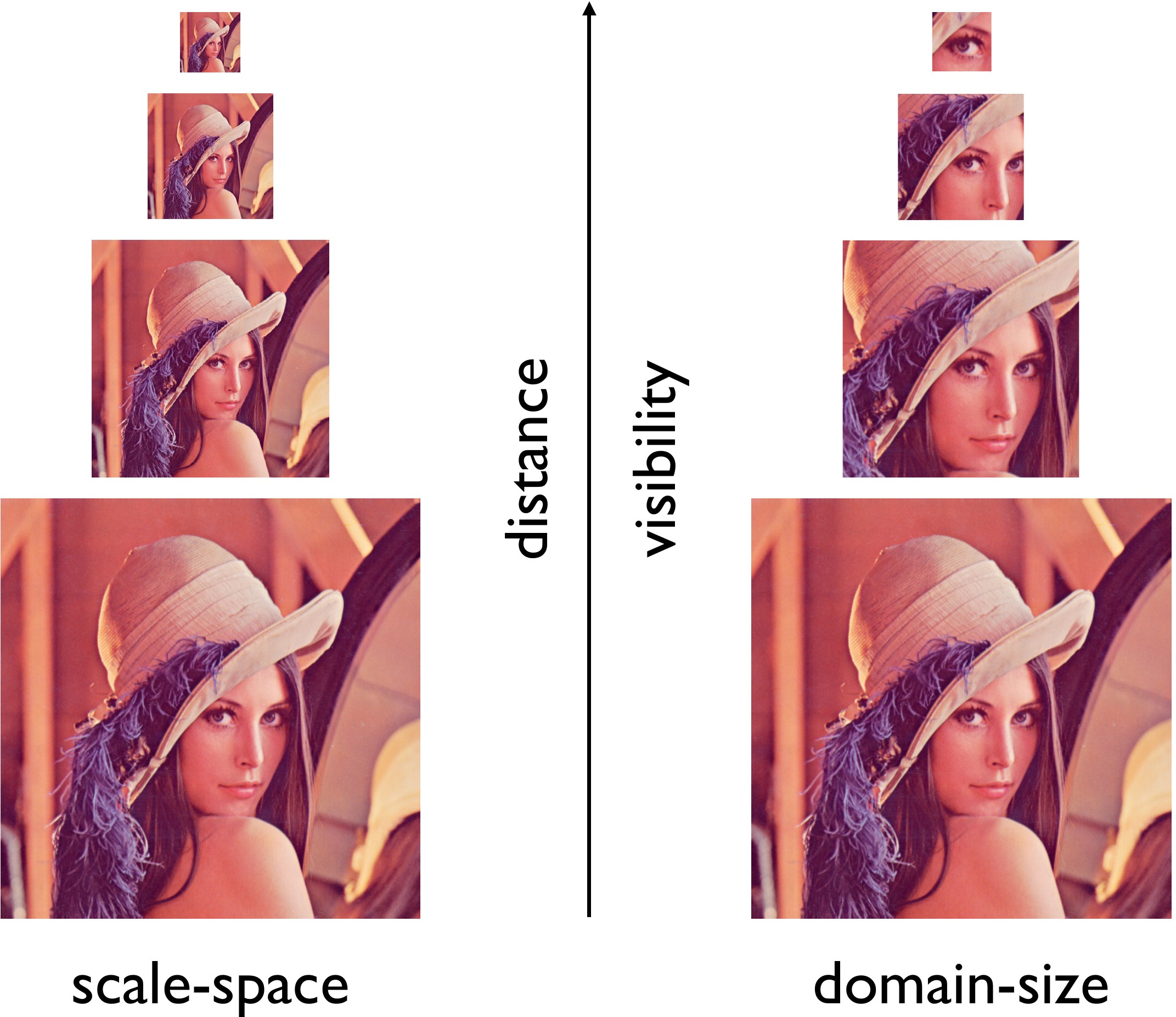}~~~
\includegraphics[width=.16\textwidth]{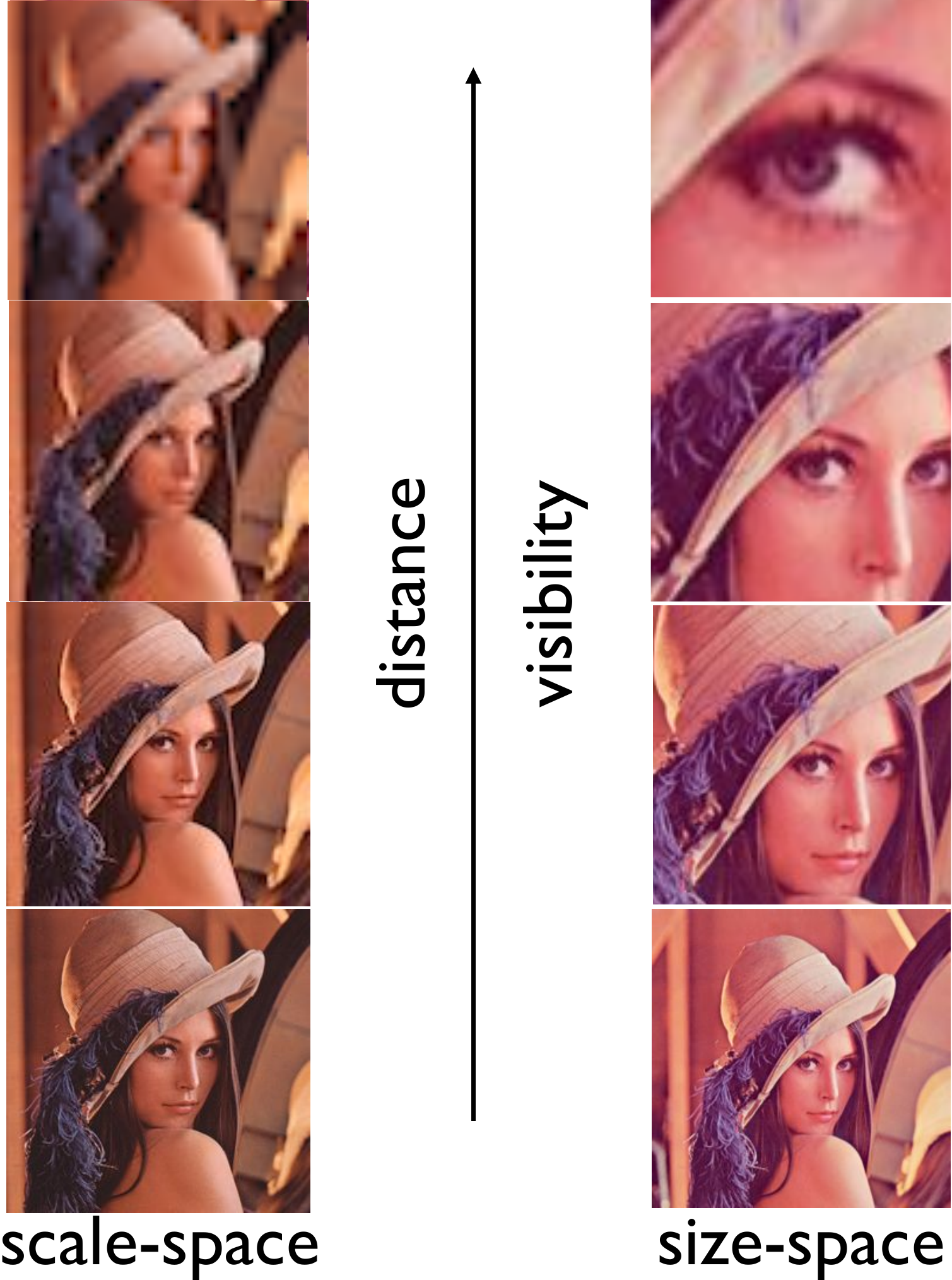}
\end{center}
\caption{\sl (Left) The ``uncertainty principle'' links the size of the domain (abscissa) of a filter to its spatial frequency (ordinate): As the data is analyzed for the purpose of compression, regions with high spatial frequency must be modeled at small scale (tall and skinny rectangle), while regions with smaller spatial frequency can be encoded at large scale (short and fat rectangle). When the task is correspondence, however, the size of the co-visible domain is independent of the spatial frequency of the {\em scene} within, as the portion that will be co-visibile in a test image is unrelated to its photometry. The resulting scale-space (right) differs from traditional scale-space, that consists of a continuum of images obtained by smoothing and downsampling a base image. The latter is relevant to searching for correspondence when the distance to the scene changes. Size-space, on the other hand, refers to a scale-space obtained by maintaining the same scale of the base image, but considering subsets of it of variable size. It is relevant to searching for correspondence in the presence of occlusions.}
\label{fig-scale-size}
\end{figure}
}
Together with the SOA likelihood, this also justifies the local marginalization of {\em domain sizes}, along with translation, as recently championed in \cite{dongS15}.

\begin{cor}[DSP-SIFT]
The DSP-SIFT descriptor \cite{dongS15} approximates an optimal representation \eqref{soa-alpha} for $G$ the group of planar similarities and local contrast transformations, when the scene is a single training image, and the test image is restricted to a subset of its domain.
\end{cor}
\cut{SIFT as designed violates the sampling principles described here, as sampling occurs with respect to the full similarity group (positions, scales and rotations are selected using a co-variant detector), but {\em anti-aliasing} is only performed in position (spatial pooling) and orientation (histogram smoothing), {\em not in scale}, which in SIFT is tied to domain size.} 
The assumptions underlying all local representations built using a single image break down when the scene is not flat and not moving parallel to the image plane. In this case, multiple views are necessary to manage nuisance due to general viewpoint changes.

\vspacetitle
\subsection{General viewpoint changes} 
\label{sect-viewpoint}
\vspacetitle

If a co-variant translation-scale {\em and size} sampling/anti-aliasing mechanism is employed, then around each sample the only residual variability to viewpoint $SE(3) = \real^3\times SO(3)$ is reduced\cut{\footnote{In reality, translation in space is not equivalent to translation and scaling of the image plane, for the former induces deformations of the image domain due to parallax effects and occlusions, which are absent in the latter. However, locally and away from occlusions, one is a first-order approximation of the other, so the derivation is valid for each local region that does not straddle an occluding boundary, justified by our handling of occlusions via the restriction to receptive fields in Sect. \ref{sect-occlusion}.} to $SO(3)$. That can be further factored into a rotation of the image plane (``in-plane'' rotation), and its complement (``out-of-plane'' rotation).\cut{ We next show how in-plane rotations can be eliminated, leaving only out-of-plane rotations}}.

\label{ex-rotation}
\cut{Canonization is the process by which a co-variant detector (a functional of the data and a chosen group whose zero-level set identifies isolated elements of the group that co-vary with it) is used to determine (multiple) local reference frames with respect to which the data is, by construction, invariant to the chosen group \cite{soatto10}. This procedure is particularly well suited to deal with planar rotation, since the statistics of natural images ensure that with high probability orientation-co-variant detectors have few isolated extrema. An example is the local maximum of the norm of the gradient along the direction $\ttheta = \hat \ttheta_l(\xx)$.\footnote{Here $g$ acts on $\xx$ via $g\xx(u_i, v_i) = \xx(u_i'', v_i'')$ where $u'' = u \cos \ttheta  -v \sin\ttheta$ and $v'' = u \sin\ttheta  + v \cos \ttheta$, and a canonical element $\hat g_l(\xx) = \hat \ttheta$ can be obtained as $\hat \ttheta = \arg\max_\ttheta \| \nabla \xx(u_i', v_i') \|$. The corresponding rotation invariant $\hat g^{-1}(\xx) \xx$ is $\angle \nabla \xx(u_i', v_i')$ where $u' = u \cos \ttheta  + v \sin\ttheta$ and $v' = - u \sin\ttheta  + v \cos \ttheta \doteq \ttheta'$.}
Invariance to $G = SO(2)$ can be achieved by retaining the samples
\be
p_{\M}(\ttheta | G) = \{ p_\M(\ttheta | \hat \ttheta_l) \}_{l = 1}^L
\ee
Rotation anti-aliasing is performed by regularizing the orientation histogram.  Note that, as it was for contrast, planar rotations can affect both the training  $\xx$ and the test image $\f$.} In some cases, a consistent reference (canonical element) for both training and test images is available when scenes or objects are geo-referenced: The projection of the gravity vector onto the image plane \cite{jonesS09}. In this case,\cut{ $L = 1$, and} $\hat \ttheta$ is the angle of the projection of gravity onto the image plane  (well defined unless they are orthogonal). Alternatively, multiple (principal) orientation references can be selected based on the norm of the directional derivative \cite{lowe04distinctive}:
\be
p_{\M}(\ttheta | G) = p_\M(\ttheta | \hat \ttheta).
\ee
%\cut{In reality, rotation canonization should contend with spatial quantization, neglected here since rotation errors are absorbed by the binning of gradient orientation $\epsilon_\ttheta$. }
This leaves out-of-plane rotations to be managed.\cut{ Unfortunately, the effects of such rotations on future images depend on the shape of the underlying scene, which is unknown, and that cannot be determined from a single image. Therefore, the only way in which true viewpoint changes can be factored out of the representation is if multiple training images {\em of the same scene} are available.} \cite{dongHBSB13} have proposed extensions of local descriptors to multiple views, based on a sampling approximation of the likelihood function, $\hat p_\M$,  or on a point estimate of the scene $p_{\hat \M}$, MV-HoG and R-HoG respectively. The estimated scene has a geometric component (shape) $\hat S$ and a photometric component (radiance) $\hat \rho$, inferred from the LA model as described in \cite{soattoD12ICVSS}\cut{. These in turn enable the approximation of the predictive likelihood $p_{\hat \M, G}$, and hence the representation: 
\be
\phi_{\hat \M, G}(\ttheta_i) = \int_{SO(3)} \mathcal{N}_{{\mathbb S}^1}(\ttheta_i - %\angle 
\hat \rho \circ g \circ \pi_{\hat S}^{-1}(u_j, v_j); \epsilon_\ttheta) \| \nabla {\hat \rho} \| {\cal N}_{\sigma}(i-j)\cut{{\cal E}_{s}(\f)} d\mu(j) dP_{SO(3)}(g)
\label{r-hog}
\ee
where $\hat \M = (\hat S, \hat \rho)$, $\angle \nabla \f = \ttheta$ and $\pi^{-1}$ is the pre-image of a perspective projection (the point of first intersection of the ray through the pixel $(u_j, v_j)$ with the surface $\hat S$). Alternatively, a sampling approximation of the likelihood function $\hat p_\M(\xx^t)$ yields ``multi-view HOG''
\be
\phi_{G}(\ttheta_i | \xx^t) \\
\doteq \frac{1}{t}\sum_{\tau} \int_{\real^2}\mathcal{N}_{{\mathbb S}^1}(\ttheta_i - \angle \nabla {\xx_\tau}_j; \epsilon_\ttheta) {\cal N}_{\sigma}(i-j)\cut{{\cal E}_{s}(\f)} d\mu(j)dP(\sigma) 
\label{mv-hog}
\ee
Note that the gradient weight $\| \nabla \xx_\tau \|$ is absent, since individual samples of past data do not enable separating nuisance from intrinsic variability, and each sample image $\xx_\tau$ has different contrast, so the factor cannot be simply eliminated by normalization as done in Rem. \ref{rem-clamp} for a single image. Therefore, in MV-HOG it is necessary to assume that training images are captured under the same illumination conditions.  In MV-HOG, regularization is implicit in the kernel, and the predictive likelihood is based on simple planar transformations. In R-HOG, the estimated scene (which requires regularization to be inferred) acts as the regularizer \cite{dongHBSB13}.
\cutForReview{Intermediate solutions where the planar affine or projective group is marginalized result in increased complexity at a modest performance gain. Solutions beyond spatial rigid motion, to arbitrary homeomorphisms of the plane, result in a loss of discriminative power, for in that case co-occurrence is the only spatial property that is preserved and one ends up with a ``bag of features'' model}}.
Once the effects of occlusions are considered (which force the representation to be local), and the effects of general viewpoint changes are accounted for (which creates the necessity for multiple training images of the same scene), a maximal contrast/viewpoint/occlusion invariant can be approximated: \cut{Using \eqref{r-hog},} the SOA likelihood \eqref{soa-alpha} becomes:
\be
\small
\hat L_{SE(3),\epsilon(N)}(\alpha_i) = \max_k  \Big\{ \int_{SO(3)} {\cal N}_{{\mathbb S}^1}(\alpha_i - \hat \rho\circ g_k g \circ \pi_{\hat S}^{-1}(\xx_j); \epsilon_\alpha) \kappa_\sigma(i-j) d\mu(j)dP(\sigma)  dP_{SO(3)}(g)) \Big\}_{k=1}^N
\label{eq-final}
\ee
in addition to domain-size pooling. The assumption that all existing multiple-view extensions of SIFT do {\em not} overcome is the conditional independence of the intensity of different pixels\cut{ \eqref{eq-prod}}. This is discussed in \cite{soattoC14ICLR} for the case of convolutional deep architectures, and in the next section for Scattering Networks.  Capturing the joint statistics of different components of the SOA likelihood is key to modeling intra-class variability of object or scene categories.

\subsection{DSP-Scattering Networks}
\label{sect-dsp-sca}

The scattering transform \cite{mallatB11} convolves an image (or patch) with a Gabor filter bank at different rotations and dilations, takes the modulus of the responses, and applies an averaging operator to yield the scattering coefficients. This is repeated to produce coefficients at different layers in a scattering network. The first layer is equivalent to SIFT \cite{mallatB11}, in the sense that \eqref{eq-contr-inv} can be implemented via convolution with a Gabor element with orientation $\ttheta$ then taking the modulus of the response.\cut{ This is the convolution-modulus step in the scattering transform. Then the local marginalization corresponds to a low-pass filter applied to the histogram, yielding a spatially-pooled (regularized) histogram of gradient orientations. The same operator exists in the scattering network where the modulus of the filter response goes through a low-pass filter to generate the final coefficients.} 
One could conjecture that domain-size pooling (DSP) applied to a scattering network would improve performance in tasks that involve changes of scale and visibility.  We call the resulting method DSP Scattering Transform (DSP-SC).  Indeed, this is the case, as we show in the Appendix of \cite{soattoDK15}, where we compare DSP-SC to the single-scale scattering transform (SC) to the datasets of \cite{mikolajc03survey} (Oxford) and \cite{fischer2014descriptor}. 

\cut{Stacked architectures, such as the SC, offer the promise to lift the strong spatial independence assumption implicit in SIFT and its variant, as well as DSP-SIFT \cite{dongS15}. In the next section we discuss an extension to deep architectures. }

\subsection{DSP-CNN}
\label{sect-DSP-CNN}

Deep convolutional architectures can be understood  as implementing successive approximations of an optimal representation by stacking layers of (conditionally) independent local representations of the form \eqref{eq-final}, which have been shown by \cite{soattoC14ICLR} to increasingly achieve invariance to large deformations, despite locally marginalizing only affine (or similarity) transformations. As \cite{dongS15} did for SIFT, and as we did for the Scattering Transform above, we conjectured that pooling over domain size would improve the performance of a convolutional network. In the Appendix of \cite{soattoDK15}, we report experiments to test the conjecture using a pre-trained network which is fine-tuned with domain-size pooling on benchmark datasets. 

\subsection{DSP-DPM}
\label{sect-DSP-DPM}

We have also developed domain-size pooling extensions of deformable part models (DPMs) \cite{felzenswalb}, small trees of local HOG descriptors (``parts''), whereby local photometry is encoded in the latter (nodes), and geometry is encoded in their position on the image relative to the root node (edges). Intra-class shape variability is captured by the posterior density of edge values, learned from samples. Photometry is captured by a ``HOG pyramid'' where the {\em size} of each part is pre-determined and fixed relative to the root.\cut{ Interpreting the photometric descriptor as a likelihood function, rather than a ``feature vector,'' helps interpreting DPM as a (factorized) posterior density, where photometry is encoded by the SOA likelihood.} One could therefore conjecture that performing anti-aliasing with respect to the size of the parts would improve performance. Experimental results, reported in the Appendix of \cite{soattoDK15}, validate the conjecture.

\cut{\vspacetitle
\section{Conclusions}
\vspacetitle
~
We have derived an expression \eqref{eq-final} for minimal sufficient statistics of past data when the test image is restricted to a neighborhood of $\f$ where $\alpha_i$ is computed, corresponding to sampled locations around $(u_k, v_k)$, with scales $\sigma$ pooled according to the prior $dP(\sigma)$ around the samples $\sigma_k$. 
~
If a {\em sufficiently exciting} training set is available, spanning variability due to out-of-plane rotations, marginalization of $SO(3)$ can be replaced by temporal averaging of the training images \eqref{mv-hog}.  The joint distribution of local descriptors can be captured by a stacked architectures, as shown in \cite{soattoC14ICLR} and illustrated for Scattering Networks, Deformable Parts Models, and general deep convolutional architectures. 
}
\begin{small}
\subsubsection*{Acknowledgments} 

We acknowledge discussions with Alessandro Chiuso, Joshua Hernandez, Arash Amini, Ying-Nian Wu, Taco Cohen, Virginia Estellers, Jonathan Balzer. Research supported by ONR N000141110863, NSF RI-1422669, and FA8650-11-1-7154.
\end{small}

{\small
%% \bibliographystyle{iclr2015}
%% %\bibliographystyle{plain}
%% \bibliography{/Users/soatto/lib/tex/self,/Users/soatto/lib/tex/total2}

}

\end{document}